\documentclass[letterpaper, 10 pt, conference]{ieeeconf}  

\IEEEoverridecommandlockouts                              

\usepackage{booktabs}
\usepackage{graphics}            
\usepackage{epsfig}              
\usepackage{times}               
\usepackage{amsmath}             
\usepackage{amssymb}             
\usepackage{multicol}            
\usepackage{multirow}            
\usepackage{float}               
\usepackage{makecell}            
\usepackage{booktabs}            
\usepackage[hidelinks]{hyperref}
\usepackage{cleveref}

\title{
\LARGE \bf
Resolving Symmetry Ambiguity in Correspondence-based Methods \\
for Instance-level Object Pose Estimation
}

\author{Yongliang Lin$^{1}$, Yongzhi Su$^{2,3}$, Sandeep Inuganti$^{2,3}$, Yan Di$^{4}$, \\ Naeem Ajilforoushan$^{1}$, Hanqing Yang$^{1}$, Yu Zhang$^{1*}$, Jason Rambach$^{3*}$  
\thanks{*Corresponding authors}
\thanks{$^{1}$College of Control Science and Engineering, Zhejiang University, China
        {\tt\small lyl1020@zju.edu.cn,zhuangyu80@zju.edu.cn}}%
\thanks{$^{2}$RPTU Kaiserslautern, Germany
        {\tt\small suyongzhi526@gmail.com, sain01@dfki.de}}%
\thanks{$^{3}$German Research Center for Artificial Intelligence (DFKI), Germany
        {\tt\small jason.rambach@dfki.de}}%
\thanks{$^{4}$Technische Universit\"{a}t M\"{u}nchen, Germany
        {\tt\small yan.di@tum.de}}%
\thanks{This work has been submitted to the IEEE for possible publication. Copyright may be transferred without notice, after which this version may no longer be accessible.}
}

\begin{document}
\maketitle
\thispagestyle{empty}
\pagestyle{empty}

\begin{abstract}
Estimating the 6D pose of an object from a single RGB image is a critical task that becomes additionally challenging when dealing with symmetric objects. Recent approaches typically establish one-to-one correspondences between image pixels and 3D object surface vertices. However, the utilization of one-to-one correspondences introduces ambiguity for symmetric objects. To address this, we propose SymCode, a symmetry-aware surface encoding that encodes the object surface vertices based on one-to-many correspondences, eliminating the problem of one-to-one correspondence ambiguity. We also introduce SymNet, a fast end-to-end network that directly regresses the 6D pose parameters without solving a PnP problem. We demonstrate faster runtime and comparable accuracy achieved by our method on the T-LESS and IC-BIN benchmarks of mostly symmetric objects. Our source code will be released upon acceptance.
\end{abstract}


%
\section{INTRODUCTION}
Our goal is to accurately estimate the 6D poses of known objects from a single RGB image. Precise pose perception is crucial for manipulation, augmented reality and autonomous driving applications ~\cite{su2019deep, su2023opa}. Unlike methods incorporating depth information~\cite{2024hipose}, RGB-only approaches offer a broader range of applications at a lower cost.

Currently, high performing frameworks~\cite{wang2021gdr, su2022zebrapose}, establish correspondences between 2D image pixels and 3D object surface vertices. A Perspective-n-Point (PnP) algorithm variant~\cite{EPnP} is then used to estimate the pose. Correspondences are typically utilized in two ways: (1) as auxiliary targets~\cite{su2022zebrapose}, with the PnP algorithm applied subsequently to estimate the pose, or (2) as intermediate variables to output pose through a network~\cite{wang2021gdr}. However, the PnP algorithm requires one-to-one mapping between 2D-3D correspondences, which poses a challenge for symmetric objects. For instance, in the case of a texture-less ball, any pixel in the image can be associated with any vertex on the object's surface. The problem becomes more pronounced when the target correspondences are specified by a single ground truth pose.

Previous works~\cite{su2022zebrapose} demonstrated that symmetrical ambiguity can be mitigated by employing powerful but computationally intensive methods like RANSAC-PnP variants to handle high outlier correspondence ratios. Other works~\cite{wang2021gdr, di2021so} train end-to-end networks to directly regress pose parameters, bypassing the PnP problem. However, ambiguity issues arise in cases of severe occlusion, leading to a significant decline in pose estimation accuracy due to a notable drop in the accuracy of correspondence prediction.

\begin{figure}[t]
        \centerline{\includegraphics[width=0.50\textwidth]{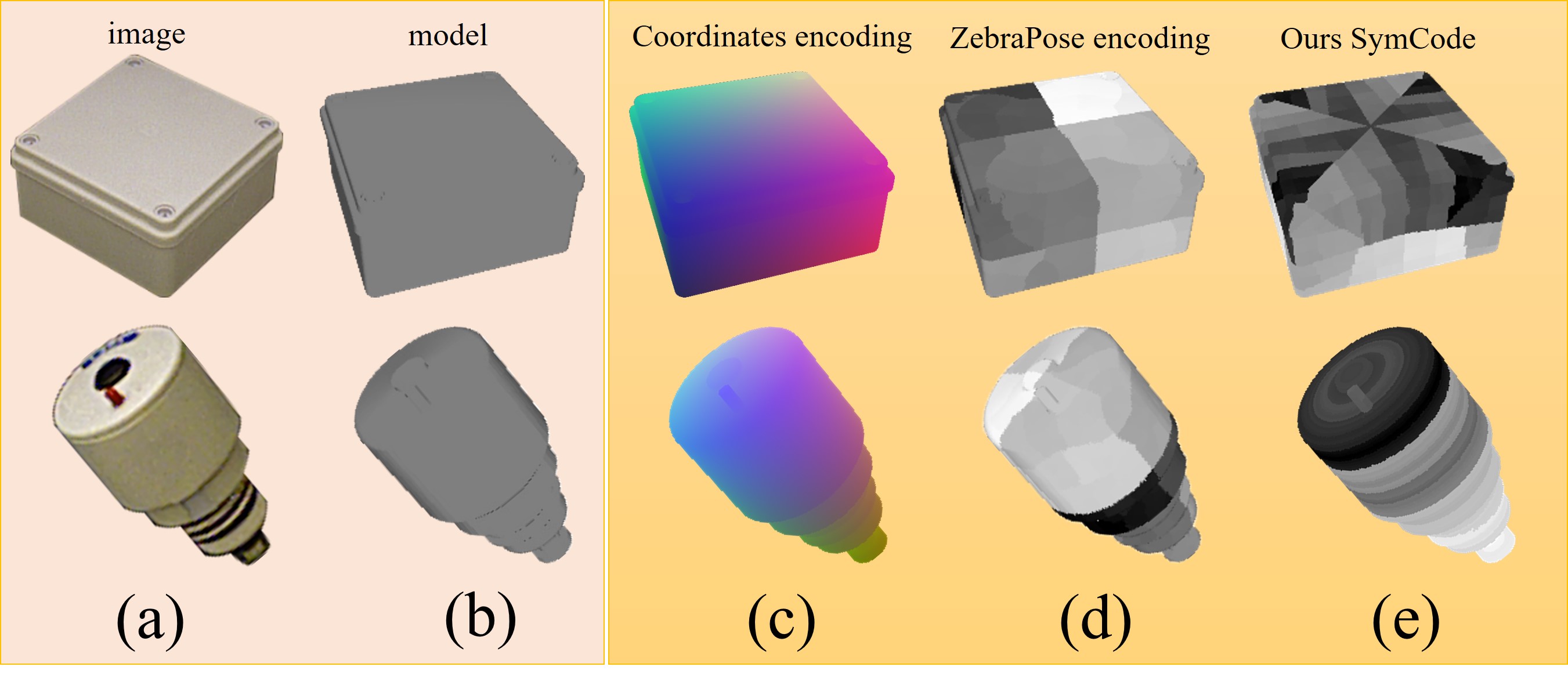}}
        \caption{\textbf{Comparison of Surface Encoding.} (a) Object images. (b) Textureless models. (c) 3D coordinate encoding. (d) ZebraPose encoding~\cite{su2022zebrapose}. (e) Our proposed SymCode. The 3D coordinate encoding and ZebraPose encoding, based on one-to-one correspondences, do not consider symmetry. In contrast, SymCode, based on one-to-many correspondences, explicitly preserves symmetry information.}
        \label{compare_encoder}
\end{figure}

To address this issue, we propose SymCode, a dense correspondence encoding that encodes one-to-many 2D-3D correspondences, benefiting not only symmetrical objects but also objects that are "almost symmetrical", meaning that they exhibit symmetry to some extent but may have slight variations or irregularities that deviate from perfect symmetry. For asymmetrical objects, the one-to-many 2D-3D correspondences converge to one-to-one 2D-3D correspondences.

\textbf{SymCode vs. ZebraPose Encoding.} We were inspired by the surface encoding introduced in ZebraPose~\cite{su2022zebrapose} to assign a discrete code to each correspondence, which facilitates the network training. SymCode, as shown in~\Cref{compare_encoder}, internally captures symmetry information but does not adhere to the one-to-one correspondence assumption of the PnP algorithm. A similar situation is faced by SurfEmb~\cite{haugaard2022surfemb}, which generates a large number of pose candidates using P3P and employs a refinement process to obtain the final pose. However, this process is time-consuming. In contrast, we introduce an end-to-end network, called SymNet, to directly output the pose from SymCode along with a segmentation mask.

To summarize, the main contributions of our work are as follows:
\begin{enumerate}
    \item We propose SymCode, a symmetry-aware binary surface encoding method that allows the estimation of one-to-many 2D-3D correspondences for pose estimation.
    \item We introduce SymNet, a network that recovers object pose from one-to-many 2D-3D correspondences without relying on the PnP-RANSAC method, leading to significant improvements in inference time.
    \item We investigate the impact of object symmetries as well as the performance of different symmetry-aware methods in correspondence-based object pose estimation through experiments on high-symmetry datasets such as T-LESS and IC-BIN.  
\end{enumerate}

\section{RELATED WORK}
Symmetry or pose ambiguity refers to the phenomenon where images of an object under different poses appear identical, making it impossible to determine the object's pose uniquely. In this section, we provide a brief overview of works that incorporate mechanisms for handling symmetries.

Pitteri~\cite{pitteri2019object} offers a general analytic approach to map symmetrical identities to the same pose, but results in discontinuity of the mapped pose, potentially causing abrupt changes in the network output.
CosyPose~\cite{labbe2020cosypose} utilizes the symmetry-aware ADD-S loss, which measures pose accuracy by considering the closest symmetric pose of the object as ground truth. ES6D~\cite{mo2022es6d} categorizes symmetry into five categories and proposes a symmetry-invariant A(M)GPD loss based on these categories, which outperforms the ADD(S) loss. However, this method cannot be directly applied to correspondence-based methods.

Recent correspondence-based methods typically learn one-to-one correspondences representation as auxiliary targets. However, symmetry can introduce correspondence ambiguity, causing pixels in the image to have non-unique corresponding vertices on the model, thereby leading to pose ambiguity.~\cite{richter2021handling} proposes a representation called "closed symmetry loop" to obtain one-to-many correspondences, where information represents the angle between different points and the object origin. This information is fused to achieve one-to-one correspondence.
EPOS~\cite{hodan2020epos} utilizes surface fragments to address symmetry and establishes one-to-many 2D-3D correspondences. However, the final pose is estimated from a sampled triplet of correspondences by the P3P solver. This approach still has some limitations in continuous symmetry for regressing 3D coordinates on fragment coordinate systems.
GDR-Net~\cite{wang2021gdr} guides pose regression by computing the loss with respect to the closest symmetric pose and incorporates surface fragments from EPOS~\cite{hodan2020epos}.
Surfemb~\cite{haugaard2022surfemb} learns one-to-many correspondences in a self-supervised manner but samples one-to-one correspondences from the one-to-many correspondences to adapt to the RANSAC-P3P algorithm input. 
To the best of our knowledge, our work is the first to directly recover pose from one-to-many correspondences. 

\section{METHOD}

This section offers a comprehensive description of our proposed one-to-many correspondence-based and symmetry-aware training method for 6D object pose estimation.

We aim to address the symmetrical ambiguity issue beyond the one-to-one correspondence approach. Drawing inspiration from the successful application of binary codes in one-to-one correspondence methods~\cite{2024hipose, su2022zebrapose}, we extend this approach to encode one-to-many correspondences, which are then fed into a network to obtain the final pose without RANSAC or the need of refinement.

\subsection{Problem Definition and Notation}

Given an RGB image with known intrinsic parameters and a set of CAD models of objects, our objective is to estimate the pose of each object (Rotation matrix $\mathbf{R} \in SO(3)$ and translation vector $\mathbf{t} \in \mathbb{R}^3$)  with respect to the camera in the image. 

There is only one ground-truth pose $\mathbf{T}=[\mathbf{R}|\mathbf{t}]$ for an asymmetric object, while a symmetric object has multiple possible ground-truth poses $\mathbf{T}_k, k=1, 2,...,n$, where $n$ is the number of ground-truth poses, which can be infinite.

ZebraPose~\cite{su2022zebrapose} constructs a binary encoding of the $N$ object model vertices, that defines binary codes $\mathbf{c}_{j}$ of $d$ bits that uniquely correspond to vertices $\mathbf{P}_j$. The binary encoding is built iteratively, by splitting the mesh into parts of equal amount of vertices at each step, and assigning a bit to each group. In the iteration $it, it \in \{0, 1, \text{\textellipsis}, d-1\}$ of the surface partition, we have $2^{it}$ separate sub-groups. Given a group of vertices, the partitioning is carried out using balanced k-means, resulting in the formation of two sub-groups. For asymmetrical objects, the one-to-many correspondences are equivalent to the one-to-one correspondences since there is no ambiguity due to symmetry. 

In the following sections, we compare one-to-one correspondences with one-to-many correspondences in detail. Subsequently, we present our approach for estimating the 6DoF object pose, which involves the entire process from surface encoding to the final pose estimation.

\subsection{2D-3D Correspondences}
\textbf{One-to-one Correspondences} Extracting one-to-one correspondences is a key component in pose estimation, which represents a match between a 2D point $\mathbf{p}_i=(u_i,v_i)$ from the observed image and a 3D point $\mathbf{P}_j=(x_j,y_j,z_j)$ from the object model, denoted as $\mathbf{o}_r = (\mathbf{p}_i, \mathbf{P}_j)$, where $\mathbf{p}_i\in \mathbb{R}^2$ and $\mathbf{P}_j\in \mathbb{R}^3$. The correspondence can be derived based on the ground-truth pose $\mathbf{T}$ point-wise:

\begin{equation}
        \mathbf{p}_i=\pi(\mathbf{T}\cdot \mathbf{P}_j) 
        \label{eq:projection}
\end{equation}

where $\pi(\cdot)$ is the projection function of a pinhole camera model using intrinsic matrix. We define a one-to-one correspondence set $\mathbf{O} = \{\mathbf{o}_1, \mathbf{o}_2, ..., \mathbf{o}_m\}$ containing $m$ one-to-one correspondences, where $\mathbf{o}_r = (\mathbf{p}_i, \mathbf{P}_j)$. The Perspective-n-Point (PnP) module aims to recover the pose $\mathbf{T}$ given a set of one-to-one correspondences. Most correspondence-based methods use the one-to-one correspondence set as an intermediate geometric representation. For an asymmetric object, the one-to-one correspondence is unique and can be recovered without ambiguity, but this is not the case for symmetric objects.

\begin{figure}[ht]
  \centerline{\includegraphics[width=0.4\textwidth]{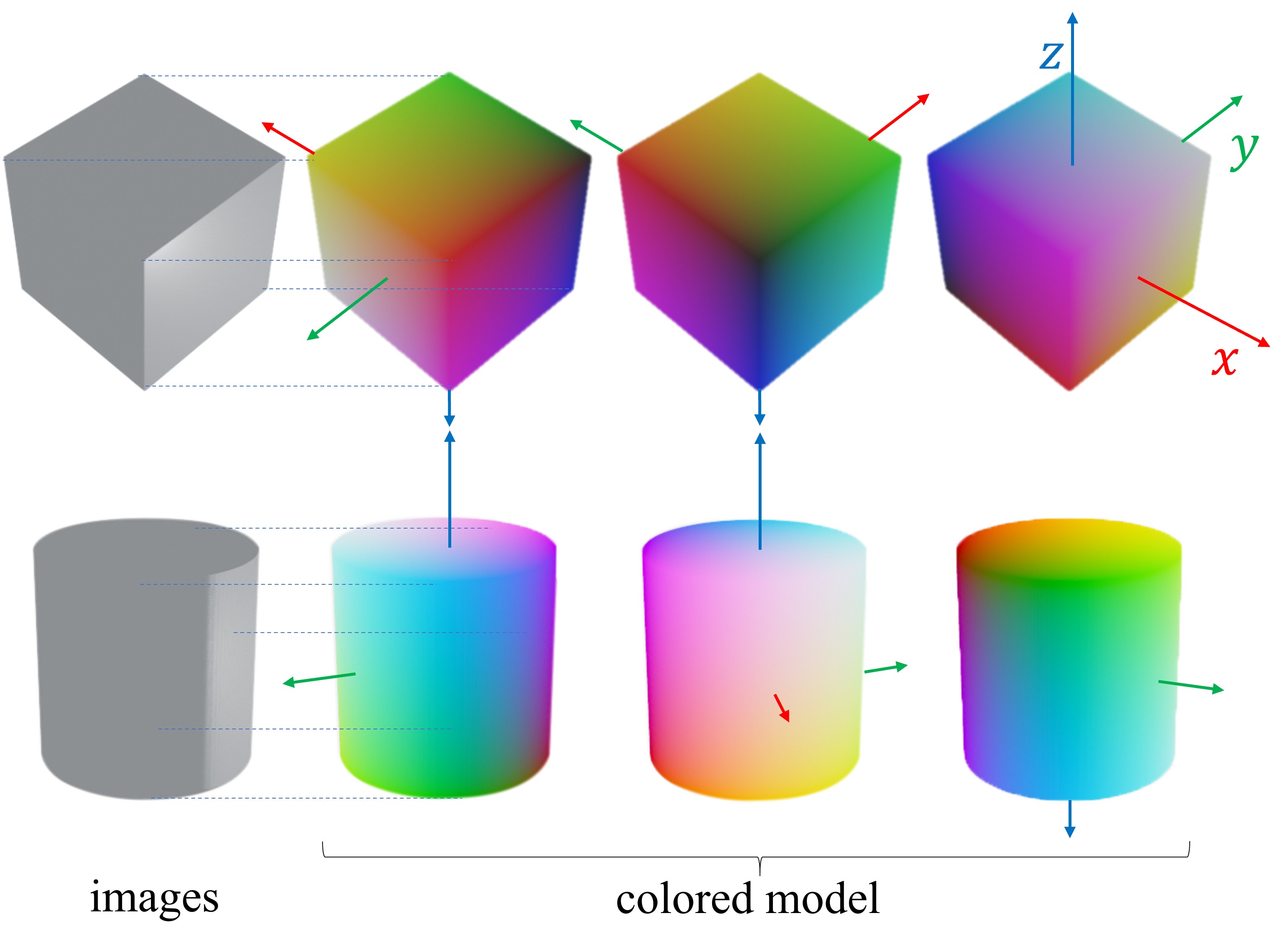}}
  \caption{\textbf{Multi possible one-to-one correspondence sets.} Left column: Images of an untextured cube and cylinder. Top: three possible correspondence sets for the cube image. Bottom: three possible correspondence sets for the cylinder image. The color of the models in the right three columns represents the coordinates in the object frame, with red, green, and blue representing the coordinates of the x-axis, y-axis, and z-axis, respectively. The object frame is represented by colored arrows as coordinates. The dashed lines show 2D-3D correspondences. Best viewed in color mode.}
  \label{fig:one_one_corres}
\end{figure}

For each ground-truth pose $\mathbf{T}_i$ of a symmetric object, we can obtain a one-to-one correspondence set using \Cref{eq:projection}. However, in the training stage, setting similar images with vastly different correspondence sets as learning targets can lead to convergence problems. We illustrate different one-to-one correspondence sets for a cube and a cylinder in \Cref{fig:one_one_corres}.

\textbf{One-to-many Correspondence.} We define a one-to-many correspondence to represent a match between a 2D point $\mathbf{p}_i=(u_i,v_i)$ and a set of all possible 3D points $\mathbf{Y}_j = \{\mathbf{P}_{j,1}, \mathbf{P}_{j,2}, ..., \mathbf{P}_{j,n}\}$, denoted as $\mathbf{o}_r = (\mathbf{p}_i, \mathbf{Y}_j)$. The correspondence should satisfy the following condition:

\begin{equation}
\mathbf{p}_i = \pi(\mathbf{T}_k \cdot \mathbf{P}_{j,k}), \quad k = 1,2,...,n
\label{eq:projection_one_to_many}
\end{equation}

\begin{figure}[ht]
\centerline{\includegraphics[width=0.4\textwidth]{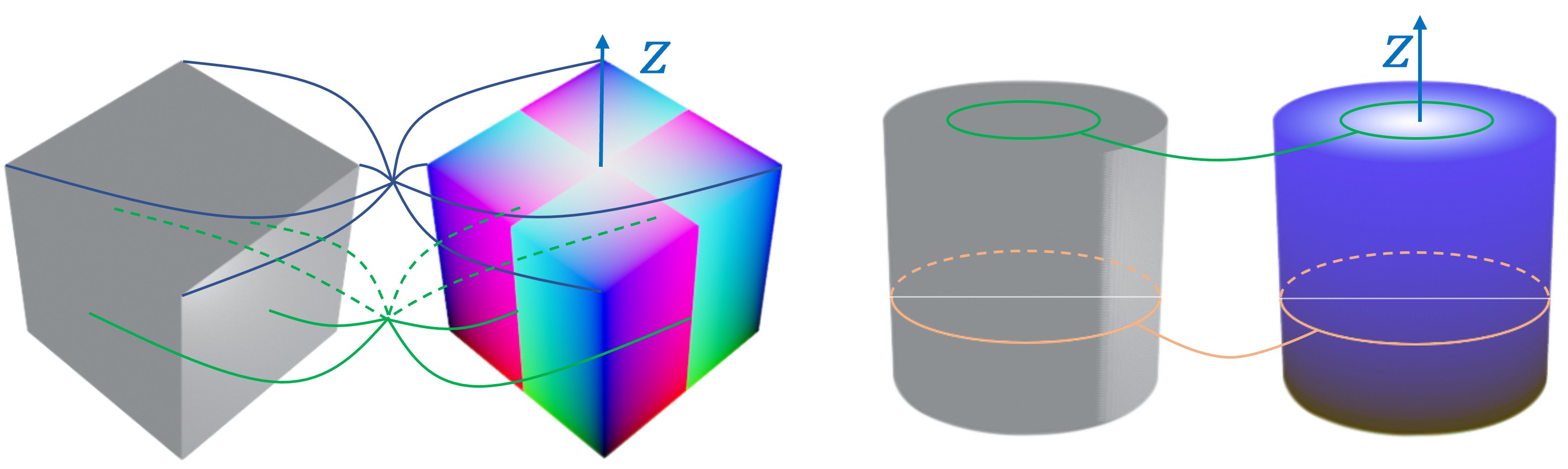}}
\caption{\textbf{One-to-many correspondences set.} All 3D points in a one-to-many correspondence are textured with the same color. Left: two specific one-to-many correspondences for corners and centers of side faces, assuming the cube has only 4 symmetries rotated along the z-axis with 0, 90, 180, 270 degrees. Invisible parts are connected by dashed lines, and visible parts are connected by solid lines. Right: two specific one-to-many correspondences for the side surface and top surface.}
\label{fig:many_many_corres}
\end{figure}

$\mathbf{T}_k$ are all the possible ground-truth poses.
We define a one-to-many correspondence set $\mathbf{O}_\text{sym} = \{\mathbf{o}_1, \mathbf{o}_2, ..., \mathbf{o}_m\}$ containing $m$ one-to-many correspondences, where $\mathbf{o}_r = (\mathbf{p}_i, \mathbf{Y}_j)$, as depicted in~\Cref{fig:many_many_corres}. 
The one-to-many correspondence demonstrate that from a pixel $\mathbf{p}_i$, it is impossible to determine the matching 3D point coordinates $\mathbf{P}_j$, but it is possible to determine the corresponding 3D point set $\mathbf{Y}_j$. The one-to-many correspondence presents an easier learning task for the network compared to the one-to-one correspondences due to its lack of ambiguity.

\begin{figure}[ht]
    \centerline{\includegraphics[width=0.5\textwidth]{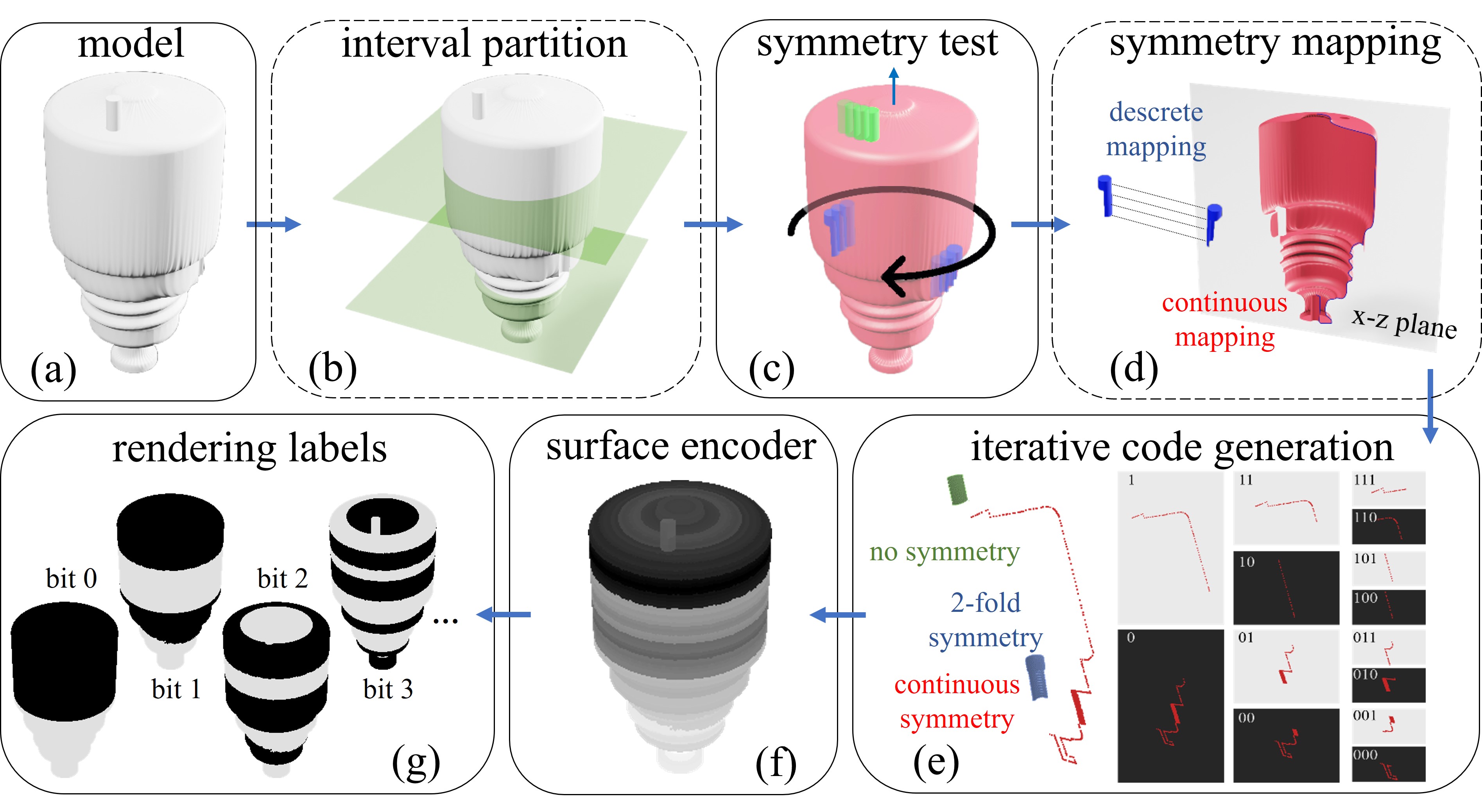}}
        \caption{\textbf{The generation process of SymCode and label rendering.} (a) Object CAD models. (b) For complex models, we manually partition the model into several parts to ensure higher accuracy in the following process. (c) Testing the symmetry type of a vertex based on symmetry priority. (d) Finding the main vertex of each one-to-many correspondence; continuous symmetry and discrete symmetry are processed differently. (e) Each correspondence is assigned a unique binary code $\mathbf{c}_i$. (f) The surface encoder encodes each vertex in the model with a binary code, which inherently preserves symmetry information. (g) The rendered label is used as an intermediate target for the network. Optional processes are enclosed in dashed boxes.}
        \label{process_of_model}
\end{figure}

\subsection{Symmetry-aware Surface Encoding}
Our objective is to encode each one-to-many correspondence $\mathbf{o}_i$ with a discrete binary code $\mathbf{c}_i$.~\Cref{process_of_model} presents a schematic overview of the proposed one-to-many surface encoding pipeline.

\textbf{Generate One-to-Many Correspondence.} When an object possesses symmetry, there are multiple rigid motions that can map one ground-truth pose to other ground-truth poses. Formally, we consider the set as follows:
\begin{equation}
\mathcal{M} = \{\mathbf{m}\in SE(3)\ |\ \mathbf{T}_1 \cdot \mathbf{m} \in \{\mathbf{T}_1, \mathbf{T}_2,..., \mathbf{T}_n\} \}
\label{eq:rigid_motions}
\end{equation}
where $\mathbf{T}_1$ represents any single ground-truth pose, $\mathbf{T}_k$ represents another ground-truth pose where $n$ denotes the total number of possible ground-truth poses.

Symmetries can be categorized into two types: discrete and continuous. In the case of discrete symmetry, the set $\mathcal{M}$ consists of a finite number of elements. For a vertex $\mathbf{P}$, we generate all other possible 3D vertices belonging to the same one-to-many correspondence as

\begin{equation}
\mathbf{P}_{j,k}=\mathbf{m} \cdot \mathbf{P},\quad k=1,2,..,n, \quad \mathbf{m} \in \mathcal{M}\label{eq:generate_correspondence}
\end{equation}


In the case of continuous symmetry, the vertices on the surface undergo rotation along the axis of symmetry, resulting in their alignment on a plane, as illustrated in ~\Cref{process_of_model}(d). By doing so, vertices that undergo rotation and end up at the same position on the plane are consolidated into the same correspondence. We can formalize the rotation process as follows:

\begin{equation}
\tilde{\mathbf{P}}=
\begin{bmatrix}
 \sqrt{x^2+y^2}  &0 &z
\end{bmatrix}^T
\label{eq:rotation_projection}
\end{equation}

Here, $\tilde{\mathbf{P}}$ is the transferred vertex of origin vertex $\mathbf{P}$, and is referred to as the main vertex. $x, y, z$ represent the three components of $\mathbf{P}$. It is assumed that the rotation occurs along the z-axis, and the resulting plane is the x-y plane, as illustrated in~\Cref{process_of_model}(d).

\textbf{Handling Mixture Symmetry.} For the T-LESS dataset, different objects are only annotated with the main type of symmetry. We can directly generate one-to-many correspondences based on this symmetry annotation. However, many objects consist of a mixture of symmetric parts. Consider an object constructed by combining a cube and a cylinder together. Although considering the entire object may suggest discrete symmetry, utilizing correspondences constructed based on discrete symmetry in occluded scenarios would still result in ambiguity.

Additionally, some objects are considered "almost symmetrical," meaning they exhibit symmetry except for a small detail. To tackle these challenges, we offer an advanced annotation tool that enables more precise labeling of object models. This is the only part of the process that is still manual, but can be completed within a few minutes only. If the object is perfectly symmetrical, this step is not necessary, making the entire annotation process fully automated, relying on symmetry annotation information from BOP~\cite{hodan2024bop}.

\Cref{process_of_model}(a) provides an example of this scenario. We explicitly categorize points on objects into the following four categories (1) no symmetry, (2) continuous symmetry, (3) discrete symmetry, and (4) n-fold symmetry. The n-fold symmetry is a special case of discrete symmetry, i.e. the angle of symmetry is 
\begin{equation}
\theta=i \cdot 2\pi /n,\quad i \in [1,...,n]
\end{equation}
along the axis. The classification is determined by the average distance between each vertex and its nearest vertex after applying the transformation in \Cref{eq:generate_correspondence}. Since different types of symmetry inherently have an inclusion relationship, we introduce the concept of symmetry priority. This priority determines the order in which we test the symmetry type of a vertex. Specifically, we follow this priority order: discrete symmetry $>$ n-fold symmetry $>$ continuous symmetry $>$ no symmetry. The primary symmetry type is provided in the dataset's meta information. However, for the details that break the symmetry, human intervention is required to specify them and ensure that the correct symmetry type is identified. Additionally, thresholds of error derived from empirical or experimental observations can also be used to assist in identifying the symmetry type $sym$. For a given symmetry type $sym$, we will generate the correspondence set $\mathcal{M}$ based on that symmetry and calculate the error under this specific symmetry type $sym$ as follows:

\begin{equation}
\mathbf{e}_{j,sym} = \sum_{m\in\mathcal{M}}\left \|  \mathbf{P}_{\text{near}} - \mathbf{m}\mathbf{P}_k \right \| 
\end{equation}
where $\mathbf{P}_{\text{near}}$ refers to the nearest vertex to the transformed vertex $\mathbf{m}\mathbf{P}_k$ on the object surface.

After removing the vertices that belong to high-priority symmetry, the remaining vertices can be classified into the low-priority symmetry category. Any vertices that are not categorized as belonging to any symmetry type will be recognized as having no symmetry.

In order to attain a more precise classification, the model can be subdivided into smaller intervals, enabling the identification of different forms of potential symmetry within each interval. This approach allows for a more detailed analysis of the symmetry structure, as depicted in \Cref{process_of_model}(b).

\textbf{Encoding Correspondence.} We have established the one-to-many correspondence set $\mathbf{O}_\text{sym}$ wherein each group $\mathbf{Y}_j$ comprises vertices denoted as $\mathbf{P}_{j,k}$. Additionally, each group $\mathbf{Y}_j$ is associated with a symmetry type $sym$. We introduce a method for encoding the one-to-many correspondence set. The object surface is encoded by incorporating a hierarchical binary grouping scheme for the one-to-many correspondence set.

The encoding indicates which corresponding group $\mathbf{Y}_j$ this pixel matches to. The encoding should ideally meet the following criteria:  1) To provide ample information for pose estimation, the encoding should exhibit significant differences for two groups $\mathbf{Y}_j$ and $\mathbf{Y}_j'$ with distinct symmetry types. 2) To facilitate easier learning for the network, correspondences that are in close proximity should have more similar encodings. 3) Each $\mathbf{Y}_j$ corresponds to a unique encoding.
In general, we select one 3D vertex, referred to as the main vertex, from each correspondence while considering the spatial proximity of neighboring correspondences. The main vertex will be utilized for implementing binary encoding in the next step.

Next, we introduce how to obtain the main vertex for each correspondence. For correspondences exhibiting continuous symmetry, we select the vertex mapped to the plane as the main vertex using~\Cref{eq:rotation_projection}. For other types of symmetry, we select the main vertex based on a simple criterion explained below, which can be replaced with alternative approaches as long as it satisfies the criteria (2). In our implementation, the used criterion calculates the sum of the absolute values of each coordinate component for each vertex in the correspondence. The main vertex, denoted as $\tilde{\mathbf{P}}$, is then chosen according to the equation:
\begin{equation}
    \tilde{\mathbf{P}} = \max_{\mathbf{P}_{j,k}}
    (\left | x_{j,k} \right | +\left | y_{j,k} \right | +\left | z_{j,k} \right |), \quad \mathbf{P}_{j,k} \in \mathbf{Y}_j
\end{equation}
As illustrated in~\Cref{process_of_model}(e), the main vertices for this object are represented by the colored vertices. Only one vertex in the correspondence is not symmetrical, which will be the main vertex, colored green.
For continuous symmetry, the main vertex, colored red, lies on the plane calculated by~\Cref{eq:rotation_projection}.
Regarding 2-fold symmetry, the vertices on the portion with higher coordinate values will serve as the main vertices, colored blue.

\textbf{Iterative code generation.} This part is inspired by the binary encoding used in the one-to-one correspondence method~\cite{su2022zebrapose}. Given the one-to-many correspondences set, we want to represent each correspondence $\mathbf{o}_i$ with a binary code $\mathbf{c}_i \in \mathbb{N}^d$, where $d$ is the length of the binary code. We construct such encodings based on the main vertex of each correspondence.

In \Cref{process_of_model}(e), we illustrate the process of performing $d$ iterations of grouping the main vertices. The grouping iteration begins with collections of all main vertices $G$. For a group $G_{i}$ with a binary code $c_i$, we divide it into two groups using k-means. The resulting groups are assigned codes $\mathbf{c}_i \ll 1$ and $(\mathbf{c}_i \ll 1) + 1$, where the operation $\ll$ denotes binary left shift. Eventually, we obtain $2^d$ groups, each with its binary code. These binary codes can be represented in decimal form from $0$ to $2^d - 1$. The surface encoder is depicted in \Cref{process_of_model}(f), where the color represents the decimal value of the binary code $\mathbf{c}_i$.

\subsection{Network Architecture}

\begin{figure*}[t]
        \centerline{\includegraphics[width=1.0\textwidth]{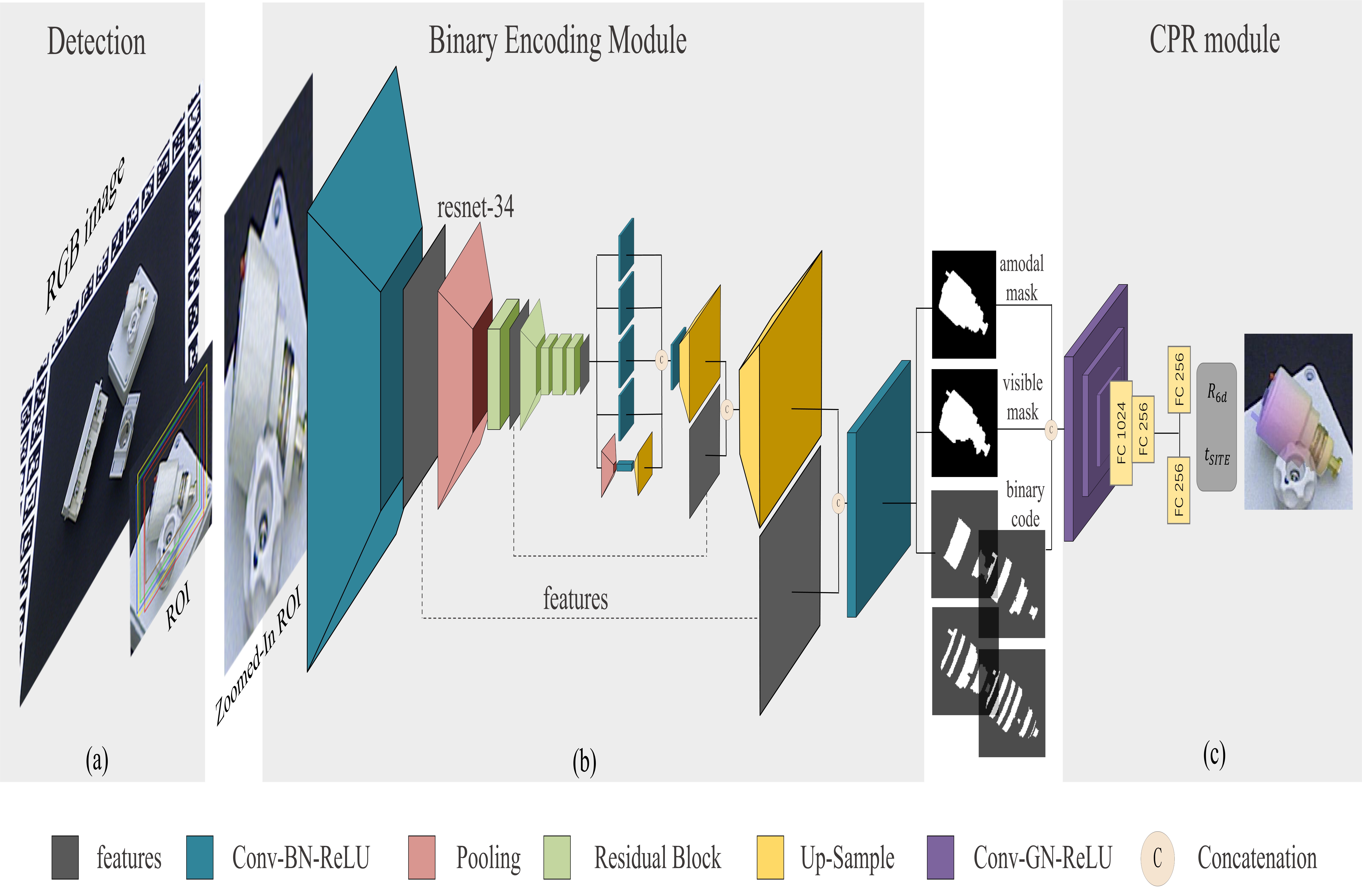}}
        \caption{\textbf{Network Architecture.} Given an RGB image, our SymNet takes the zoomed-in Region of Interest (RoI) as input and predicts intermediate features, including masks and binary code maps. The CPR module then directly regresses the 6D object pose. The entire process is an end-to-end procedure, eliminating the need for refinement or RANSAC processes for PnP.}
        \label{fig:network}
\end{figure*}

In \Cref{fig:network}, we provide the Binary Encoding Module with a zoomed-in region of interest (RoI). The outputs, namely the amodal mask $M_{\text{amo}}$, visible mask $M_\text{{vis}}$, and SymCode maps $M_{code}$, serve as intermediate variables. Subsequently, these intermediate variables are utilized as inputs to the Correspondence-based Pose Regression (CPR) module.

Our network is inspired by ZebraPose~\cite{su2022zebrapose} and GDR-Net~\cite{wang2021gdr}. We utilize a straightforward Correspondence-based Pose Regression (CPR) module to directly regress the 6D pose from the visible mask $M_\text{{vis}}$, amodal mask $M_{\text{amo}}$, and SymCode maps $M_\text{code}$. The CPR module comprises three convolutional layers, each followed by Group Normalization and ReLU activation. Subsequently, two fully connected layers are applied to the flattened features. Finally, two parallel fully connected layers output the 3D rotation parameterized as $\mathbf{R}_{6d}$~\cite{zhou2019continuity} and the 3D translation parameterized as $\mathbf{t}_\text{SITE}$~\cite{li2019cdpn}, respectively.

We utilize $L1$ loss for both the visible mask, amodal mask and SymCode maps. Our training loss is defined as
\begin{equation}
    Loss = L_\text{masks} + L_\text{code} + L_\text{params} + L_\text{ADD-S}
\end{equation}
$L_\text{params}$ corresponds to the loss for end-to-end training utilizes the $L1$ loss for $\mathbf{R}_{6d}$ and $\mathbf{t}_\text{SITE}$. Additionally, the term $L_\text{ADD-S}$ is derived from the work of~\cite{xiang2018posecnn}.

\section{EXPERIMENTS}

In this section, we introduce evaluation metrics and briefly introduce a symmetry-aware version of~\cite{su2022zebrapose} - ZebraPoseSAT, then compare our method with other model-based approaches on the T-LESS~\cite{2017tless} and IC-BIN~\cite{icbin} datasets. These datasets encompass a variety of symmetric objects, while a large synthetic-to-real domain gap exists because of texture mismatch. This makes T-LESS and IC-BIN particularly suitable for evaluating the performance of methods on symmetric objects. Furthermore, we present ablation experiments on various hyperparameters and visualizations to intuitively demonstrate the impact of SymCode.

\subsection{Evaluation Protocol}

The Visible Surface Discrepancy (VSD) evaluates the proportion of visible pixels for which the depth absolute discrepancy falls below a threshold of $\tau=20mm$. We report the recall of correct 6D object poses at $e_{VSD}<0.3$~\cite{pitteri2019object}. We also adhere to the evaluation protocol outlined in the BOP challenge~\cite{hodan2024bop} using three metrics: Visible Surface Discrepancy (VSD), Maximum Symmetry-aware Surface Distance (MSSD), and Maximum Symmetry-aware Projection Distance (MSPD). 

\subsection{ZebraPoseSAT: Alternative Symmetry-aware Solution}
To provide an symmetry-aware version of ZebraPose~\cite{su2022zebrapose} for comparison, we additionally implement ZebraPoseSAT (SAT standing for Symmetry-Aware Training), which utilizes an analytical approach~\cite{pitteri2019object} to map all ground truth poses $\mathbf{T}_i$ to a unique $\mathbf{T}$ based on the Froebenius norm, prior to generating ZebraPose Encoding. 
ZebraPoseSAT emerged as the winner of the Best RGB-Only Method in the BOP 2023 challenge~\cite{hodan2024bop}, providing a strong comparison baseline for SymNet.


\subsection{Comparison to State of the Art}

\begin{table*}[t]
        \centering
        \caption{
                T-LESS: Object recall for $err_{vsd} < 0.3$ on all Primesense test scenes. The results for the $30$ objects are grouped based on their symmetry type.
        }
        \begin{tabular}{r| c c c c c c c c}
        \toprule
        Method& \multicolumn{2}{c}{AAE \cite{sundermeyer2018implicit}} & Pix2Pose \cite{park2019pix2pose}& EdgeEnhance \cite{wen2020edge}& Pitteri\cite{pitteri2019object} & CosyPose\cite{labbe2020cosypose}& Ours(pbr) & Ours(pbr+real)\\
        \cmidrule(lr){2-3}\cmidrule(lr){4-4}\cmidrule(lr){5-5}\cmidrule(lr){6-6}\cmidrule(lr){7-7}\cmidrule(lr){8-9}
        Detector&\ \ \ \ SSD\ \ \ \ \  &\ Retina\ &\ \ Retina\ \ &\ \ Retina\ \ & Faster-RCNN & Retina & FCOS(pbr) & FCOS(pbr+real)\\
        Symmetry type& RGB & RGB & RGB &RGB& RGB &RGB-D& RGB &RGB \\
        \midrule
        Asymmetry (3)&25.98&16.95&24.73&28.76&36.533&-&66.36&\textbf{69.47}\\
        Continuous (11)&11.90&17.98&36.27&39.05&46.65&-&61.52&\textbf{63.72}\\
        Discrete (16)&14.45&18.86&25.79&32.77&38.47&-&69.51&\textbf{78.02}\\
        \midrule
        Mean&14.67 &18.35 &29.5&34.67&41.27 &62.6&66.27&\textbf{71.92} \\
        \bottomrule
        \end{tabular}
\label{tab:tless_vsd}
\end{table*}

\textbf{Results on T-LESS.} Symmetry-related works generally report the $e_{VSD}<0.3$ recall score in the T-LESS dataset. In \Cref{tab:tless_vsd}, SymNet achieves an impressive improvement of $25.0\%$ over the results reported by Pitteri~\cite{pitteri2019object}. As the Retina detector code is outdated and the detection results are not publicly available, we made use of the FCOS~\cite{fcos} detector as an alternative. For pair comparison, we also included results from CosyPose~\cite{labbe2020cosypose}, which also provides BOP scores and $e_{VSD}<0.3$ recall simultaneously. The objects of T-LESS are grouped in 3 categories based on their symmetry type and we provide the average scores for all objects in each category.

\begin{table*}[t]
        \centering
        \caption{
                BOP results on dataset T-LESS. The time is the runtime per image averaged over the dataset.
        }
        \begin{tabular}{l c c c c c c c c}
        \toprule
        6D object pose estimation method &Input type&Training type&$AR$&$AR_{VSD}$&$AR_{MSSD}$&$AR_{MSPD}$&Time(s)\\       
        \midrule
        CDPNv2~\cite{li2019cdpn}&RGB&pbr&0.407&0.303&0.338&0.579&1.849 \\
        CosyPose~\cite{labbe2020cosypose}&RGB&pbr&0.640&0.571&0.589&0.761&0.493\\
        EPOS~\cite{hodan2020epos}&RGB&pbr&0.467&0.380&0.403&0.619&1.992 \\
        ZebraPose~\cite{su2022zebrapose}&RGB&pbr&0.677&0.597&0.636&0.466&0.25 \\
        ZebraPoseSAT-EffnetB4&RGB&pbr&0.723&0.659&\textbf{0.695}&0.817&0.25 \\
        SurfEmb~\cite{haugaard2022surfemb}&RGB&pbr&0.735&\textbf{0.661}&0.686&0.857&9.043 \\
        SymNet(Ours)&RGB&pbr&\textbf{0.736}&0.631&0.693&\textbf{0.883}&\textbf{0.093} \\
        \midrule
        DPODv2~\cite{shugurov2021dpodv2}&RGB-D&pbr&0.699&0.646&0.716&0.736&0.320 \\
        \midrule
        ZebraPose~\cite{su2022zebrapose}&RGB&real+pbr&0.775&0.696&0.740&0.889&0.25 \\
        SymNet(Ours)&RGB&real+pbr&0.767&0.674&0.739&0.883&0.058\\
        \bottomrule
        \end{tabular}
\label{tab:tless_bop}
\end{table*}

\textbf{Results on BOP Benchmark.} 

We compare our results to other methods that are fully trained on synthetic data, as shown in \Cref{tab:tless_bop} and \Cref{tab:icbin_bop}. We use the default detections provided by BOP challenge 2023~\cite{hodan2024bop}. Since our method does not rely on a time-consuming pose refinement step and directly obtains the pose for every detection, our runtime is significantly reduced compared to the other methods. Our method achieves excellent results in terms of both accuracy and runtime. Specifically, our results match the accuracy of ZebraposeSAT-EffnetB4 but with a smaller backbone, and only at one third of the runtime. 

\begin{table*}[t]
        \centering
        \caption{
                BOP results on dataset IC-BIN~\cite{icbin}. The time is the runtime per image averaged over the dataset.
        }
        \begin{tabular}{l c c c c c c c c}
        \toprule
        6D object pose estimation method &Input type&Training type&$AR$&$AR_{VSD}$&$AR_{MSSD}$&$AR_{MSPD}$&Time(s)\\
        \midrule
        CRT-6D~\cite{castro2023crt}&RGB&pbr&0.537&\textbf{0.477}&0.517&0.618&0.120 \\
        ZebraPoseSAT-EffnetB4&RGB&pbr&0.545&0.475&\textbf{0.535}&0.625&0.25 \\
        SymNet(Ours)&RGB&pbr&\textbf{0.547}&0.450&0.511&\textbf{0.678}&\textbf{0.088} \\
        \bottomrule
        \end{tabular}
\label{tab:icbin_bop}
\end{table*}

\subsection{Ablation Studies}
\begin{table}[h]
        \centering
        \caption{
                Compare CPR module with PnP module on dataset T-LESS.
        }
        \begin{tabular}{c c c c c c}
        \toprule
        CPR & EPnP & $AR$&$AR_{VSD}$&$AR_{MSSD}$&$AR_{MSPD}$\\
        \midrule
                   & \checkmark & 0.283&0.166&0.190&0.493 \\
        \checkmark &            & 0.736&0.631&0.693&0.883 \\
        \bottomrule
        \end{tabular}
\label{tab:ablation_pnp}
\end{table}

\textbf{Compare with PnP solver.} In \Cref{tab:ablation_pnp}, we demonstrate the effectiveness of our CPR module. We compare the performance of our CPR module with that of RANSAC/PnP~\cite{EPnP}. The results reveal that using PnP alone is challenging in terms of obtaining accurate correspondences, which are crucial for achieving reliable pose estimation. On the other hand, our CPR module is capable of accurately calculating the final pose.

\textbf{Compare with ZebraPose.} Due to the limitation of one-to-one correspondence, we initially assumed that ZebraPose would not perform well on symmetrical objects. However, in the BOP benchmark, we observe that ZebraPose achieves a satisfactory average recall score of 0.775, which is higher than our result of 0.767 when trained with real data. We attribute this to the utilization of real images during training. Since most objects are not perfectly symmetrical, the network trained with real data can capture subtle variations and mitigate the symmetry problem. The results we can access are specifically for the model with a larger backbone, referred to as ZebraPoseSAT-EffnetB4, which has been trained solely on PBR data. However, the results for the ZebraPose model trained exclusively on PBR data are not publicly available. We performed the experiment ourselves using open-source project code, and the average recall score is $0.677$. In the same setting, our score $0.736$ exhibits improvements in both accuracy and inference time.

\begin{figure}[th]
        \centerline{\includegraphics[width=0.45\textwidth]{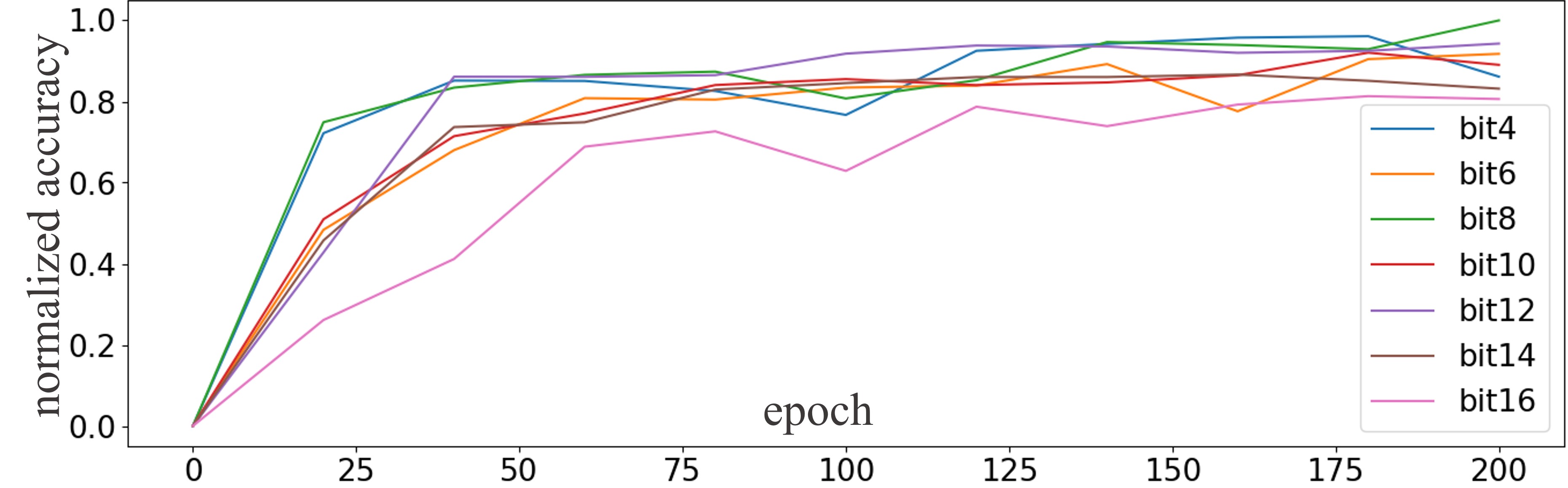}}
        \caption{We performed an ablation study on object 27 from T-LESS dataset, which exhibits discrete symmetry, to determine the optimal length of the binary code $d$. We normalized the BOP score to represent accuracy. The resulting curve demonstrates that even with a small number of bits (8 bits), our method is capable of capturing the object's pose.}
        \label{fig:ablation_bit}
\end{figure}

\textbf{Length of SymCode.} The default setting of length for $d$ is $16$ as used in ZebraPose~\cite{su2022zebrapose}. We investigate the impact of varying $d$ on the training process, as illustrated in~\Cref{fig:ablation_bit}. The results indicate that our approach is robust to variations in the length of the binary code.
Indeed, the accuracy achieved with different values of $d$ can be unpredictable, and there is no clear way to determine an optimized value beforehand. As a result, in our main results, we have chosen to fix the length of the binary code to the default value of $16$. 

\begin{figure}[ht]
    \centerline{\includegraphics[width=0.5\textwidth]{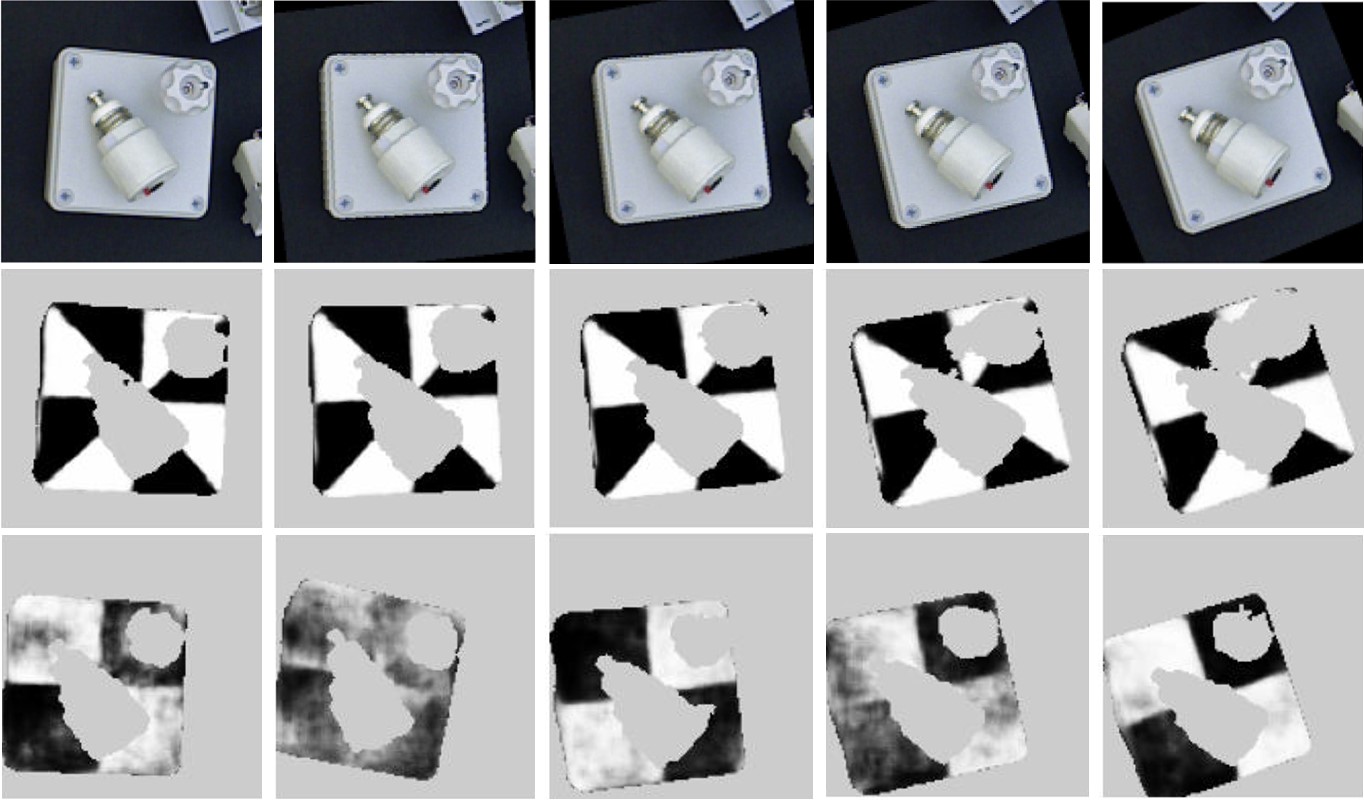}}
    \caption{\textbf{The challenge of learning one-to-one correspondences.} First line: The images with similar viewpoints. Second line: the 2nd bit prediction of SymCode. Third line: the 2nd bit prediction of ZebraPose Code. In order to achieve high accuracy pose estimation, ZebraPose Code relies on RANSAC-PnP to filter out inconsistent correspondences. However, there are still cases (second column) where ZebraPose struggles due to the ambiguity arising from symmetries. Please note that the result was obtained using our SymNet network trained with ZebraPose Code. To achieve a clear visualization, we have applied a mask based on the predicted mask to remove the background. In the visualization of the predictions, we have represented a value of 0 as black and a value of 1 as white. Any value between 0 and 1 is displayed as a shade of gray. }
    \label{fig:difficult_learn_one_to_one}
\end{figure}

\textbf{Challenge of learning one-to-one correspondences.} We conducted experiments involving training our end-to-end network using ZebraPose encodings, which is an efficient way to encode one-to-one correspondences. In this case, the network achieved a BOP recall of 0.612, which is relatively weaker compared to SymNet's performance of 0.736. When we removed all the end-to-end loss and focused solely on training the ZebraPose encodings, \Cref{fig:difficult_learn_one_to_one} illustrates the challenges associated with learning one-to-one correspondence-based encodings.

\begin{figure}[th]
    \centerline{\includegraphics[width=0.5\textwidth]{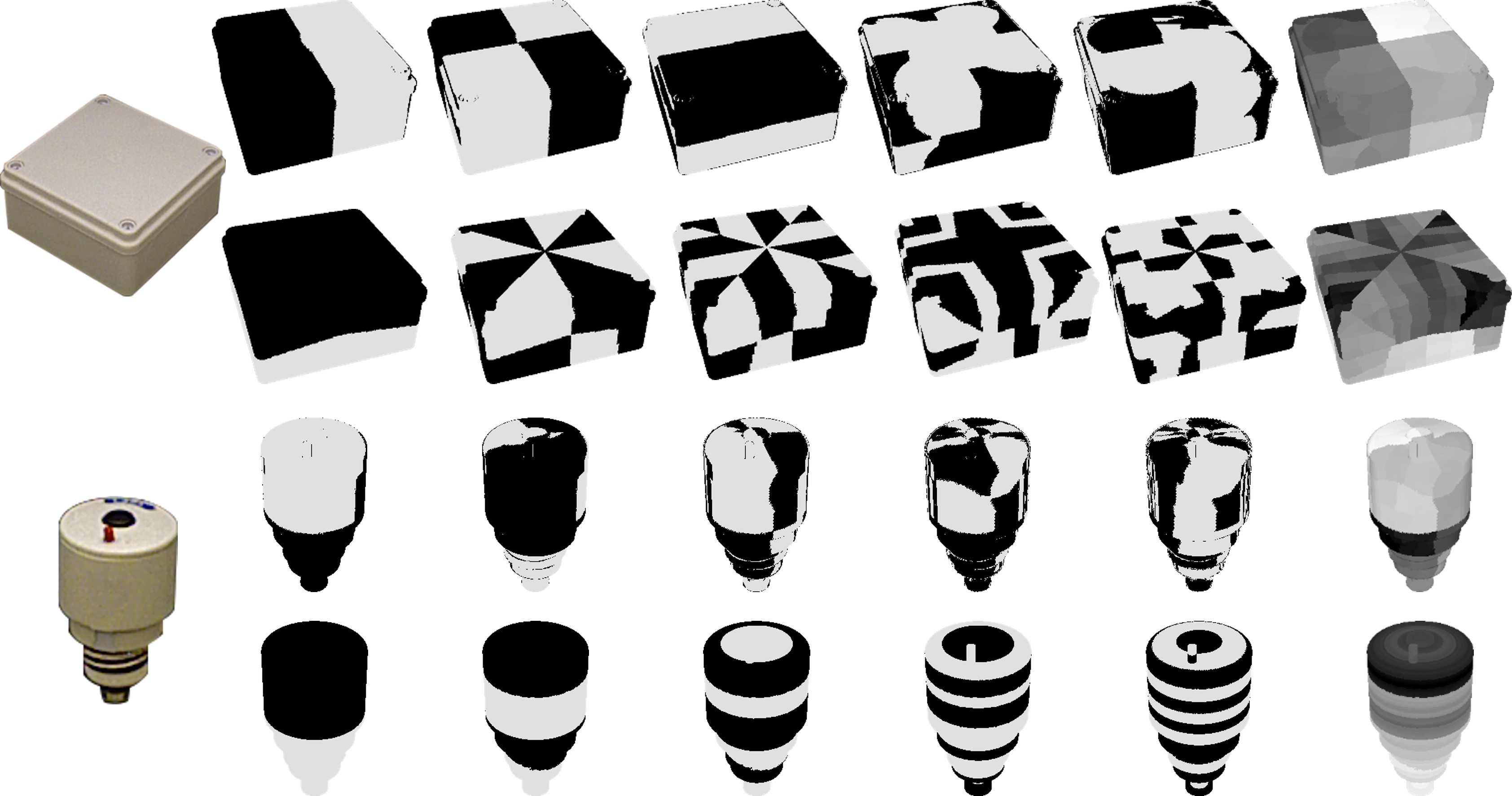}}
    \caption{\textbf{Comparison of code maps.} Detailed binary code maps representing SymCode and ZebraPose are visualized for two objects. The first and third rows depict the code maps for ZebraPose, while the second and fourth rows illustrate those for SymCode. The final columns aggregate the results for all the bits.}
    \label{fig:compare_code_map}
\end{figure}

\begin{figure}[ht]
    \centerline{\includegraphics[width=0.5\textwidth]{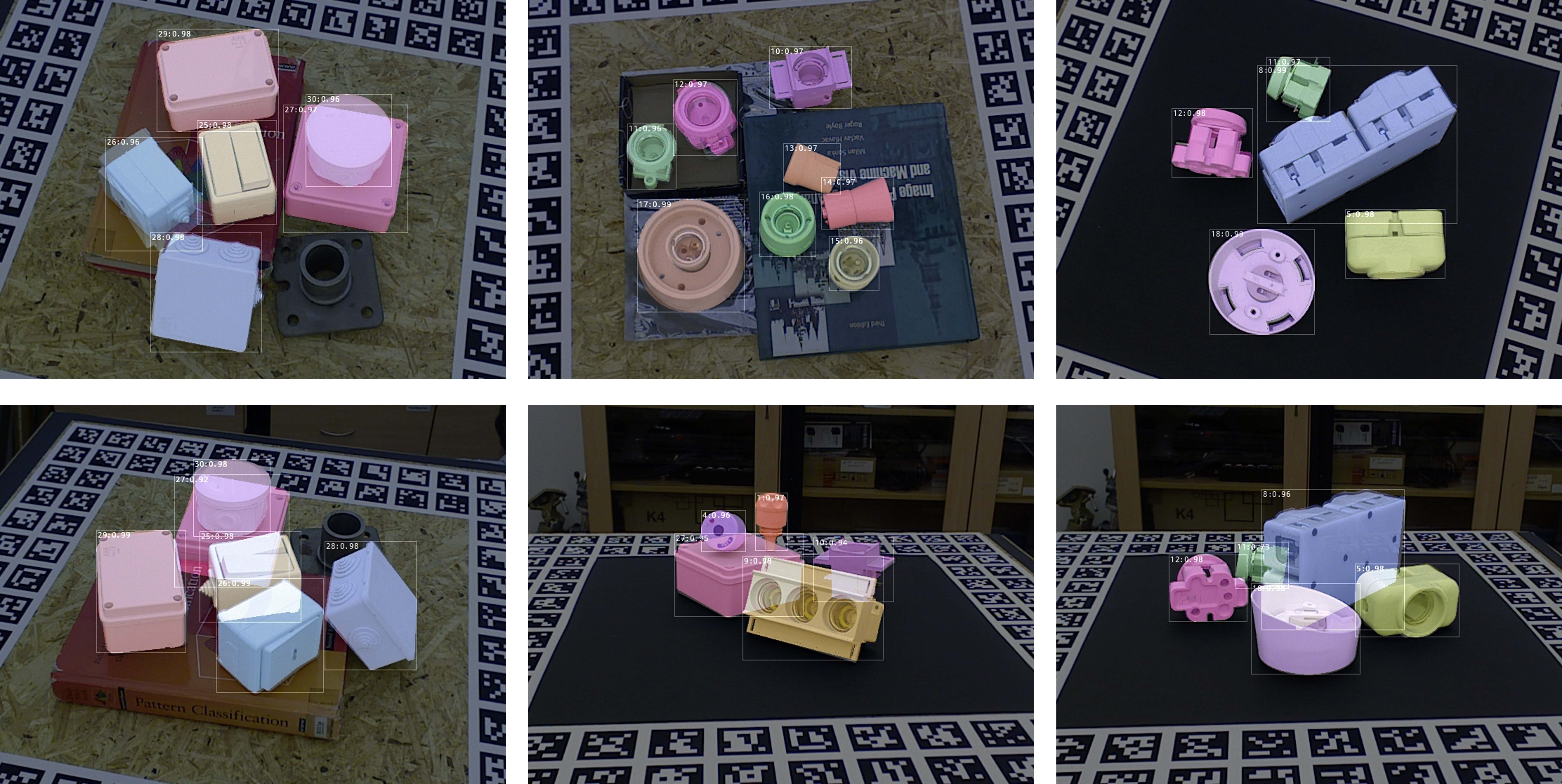}}
    \caption{\textbf{Qualitative Result on T-LESS~\cite{2017tless}.} We render the objects with the estimated pose on top of the original images. The presented confidence score are from the 2D object detection provided by BOP challenge 2023~\cite{hodan2024bop}. Each column represents a scene captured from a different viewpoint. The visualizations demonstrate that our method is capable of effectively handling severe occlusion in clustered environments.}
    \label{fig:visualization_tless}
\end{figure}

\begin{figure}[th]
        \centerline{\includegraphics[width=0.5\textwidth]{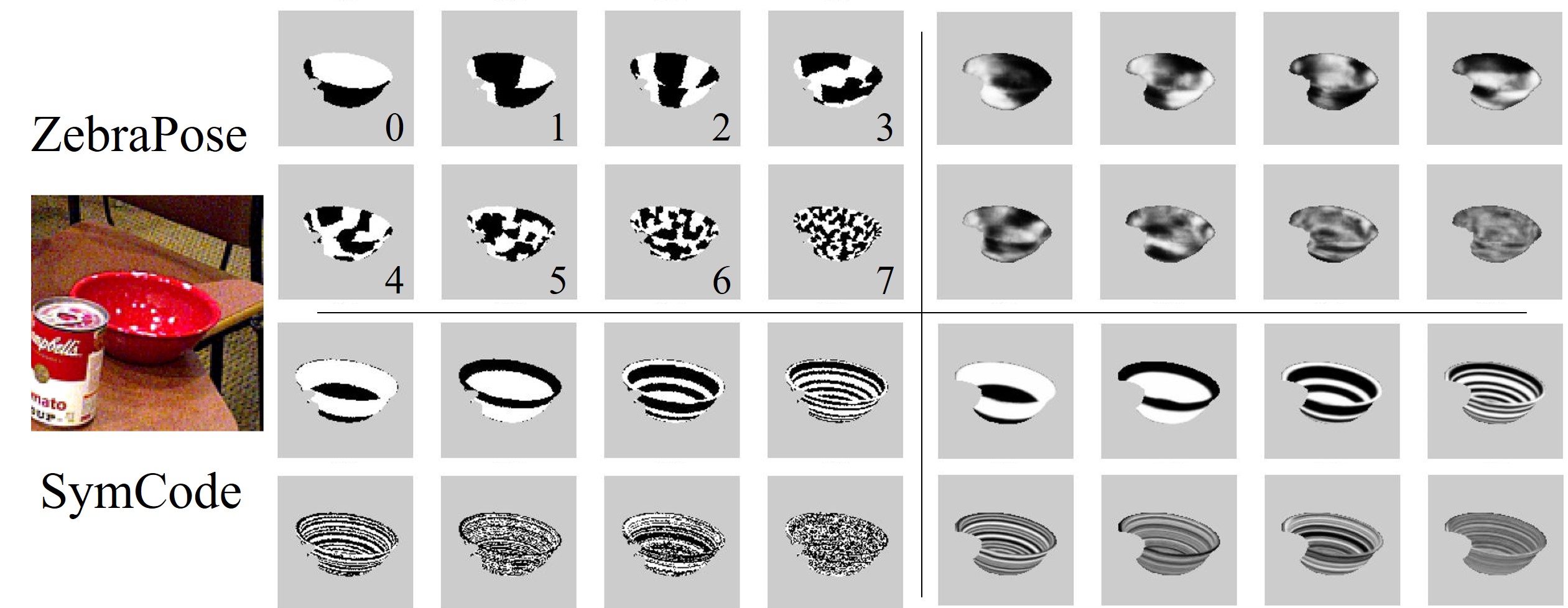}}
        \caption{\textbf{Compare with ZebraPose.} Upper Left: Ground truth of ZebraPose. Upper Right: Predicted output of ZebraPose. Bottom Left: Ground truth of SymCode. Bottom Right: Predicted output of SymCode. Only the first 8 bits are displayed. It appears that Symnet tends to be more confident in its outputs, often producing predictions closer to extreme values of 0 or 1. On the other hand, ZebraPose's predictions tend to be more centered around 0.5, indicating a lower level of confidence.}
        \label{fig:compare_zebrapose}
\end{figure}

\textbf{Visualization.} Detailed binary code maps representing SymNet and ZebraPose are visualized in~\Cref{fig:compare_code_map}, which shows our binary codes contain symmetry information. Qualitative results on T-LESS~\cite{2017tless} can be found in \Cref{fig:visualization_tless}. A visual representation of the ground truth and predicted code maps for ZebraPose and Symnet show in~\Cref{fig:compare_zebrapose}. From the visualization, it is evident that ZebraPose struggles to accurately reproduce the ground truth, whereas Symnet exhibits better performance in this regard.

\section{CONCLUSION}
Our method incorporates a symmetry discrete surface encoding technique to effectively handle symmetries. Furthermore, we demonstrate the ability to recover the pose from one-to-many 2D-3D correspondences. This approach holds the potential to influence not only other correspondence-based methods but also various other fields related to pose estimation.





\bibliographystyle{IEEEtran}
\bibliography{ref}

\begin{thebibliography}{10}
\providecommand{\url}[1]{#1}
\csname url@rmstyle\endcsname
\providecommand{\newblock}{\relax}
\providecommand{\bibinfo}[2]{#2}
\providecommand\BIBentrySTDinterwordspacing{\spaceskip=0pt\relax}
\providecommand\BIBentryALTinterwordstretchfactor{4}
\providecommand\BIBentryALTinterwordspacing{\spaceskip=\fontdimen2\font plus
\BIBentryALTinterwordstretchfactor\fontdimen3\font minus \fontdimen4\font\relax}
\providecommand\BIBforeignlanguage[2]{{%
\expandafter\ifx\csname l@#1\endcsname\relax
\typeout{** WARNING: IEEEtran.bst: No hyphenation pattern has been}%
\typeout{** loaded for the language `#1'. Using the pattern for}%
\typeout{** the default language instead.}%
\else
\language=\csname l@#1\endcsname
\fi
#2}}

\bibitem{su2019deep}
Y.~Su, J.~Rambach, N.~Minaskan, P.~Lesur, A.~Pagani, and D.~Stricker, ``Deep multi-state object pose estimation for augmented reality assembly,'' in \emph{2019 IEEE International Symposium on Mixed and Augmented Reality Adjunct (ISMAR-Adjunct)}.\hskip 1em plus 0.5em minus 0.4em\relax IEEE, 2019, pp. 222--227.

\bibitem{su2023opa}
Y.~Su, Y.~Di, G.~Zhai, F.~Manhardt, J.~Rambach, B.~Busam, D.~Stricker, and F.~Tombari, ``Opa-3d: Occlusion-aware pixel-wise aggregation for monocular 3d object detection,'' \emph{IEEE Robotics and Automation Letters}, vol.~8, no.~3, pp. 1327--1334, 2023.

\bibitem{2024hipose}
Y.~Lin, Y.~Su, P.~Nathan, S.~Inuganti, Y.~Di, M.~Sundermeyer, F.~Manhardt, D.~Stricker, J.~Rambach, and Y.~Zhang, ``Hipose: Hierarchical binary surface encoding and correspondence pruning for {RGB}-d 6dof object pose estimation,'' in \emph{Conference on Computer Vision and Pattern Recognition 2024}, 2024.

\bibitem{wang2021gdr}
G.~Wang, F.~Manhardt, F.~Tombari, and X.~Ji, ``Gdr-net: Geometry-guided direct regression network for monocular 6d object pose estimation,'' in \emph{Proceedings of the IEEE/CVF Conference on Computer Vision and Pattern Recognition}, 2021, pp. 16\,611--16\,621.

\bibitem{su2022zebrapose}
Y.~Su, M.~Saleh, T.~Fetzer, J.~Rambach, N.~Navab, B.~Busam, D.~Stricker, and F.~Tombari, ``Zebrapose: Coarse to fine surface encoding for 6dof object pose estimation,'' in \emph{Proceedings of the IEEE/CVF Conference on Computer Vision and Pattern Recognition}, 2022, pp. 6738--6748.

\bibitem{EPnP}
Vincent, Lepetit, Francesc, Moreno-NoguerPascal, and Fua, ``Epnp: An accurate o(n) solution to the pnp problem,'' \emph{International Journal of Computer Vision}, 2009.

\bibitem{di2021so}
Y.~Di, F.~Manhardt, G.~Wang, X.~Ji, N.~Navab, and F.~Tombari, ``So-pose: Exploiting self-occlusion for direct 6d pose estimation,'' in \emph{Proceedings of the IEEE/CVF International Conference on Computer Vision}, 2021, pp. 12\,396--12\,405.

\bibitem{haugaard2022surfemb}
R.~L. Haugaard and A.~G. Buch, ``Surfemb: Dense and continuous correspondence distributions for object pose estimation with learnt surface embeddings,'' in \emph{Proceedings of the IEEE/CVF Conference on Computer Vision and Pattern Recognition}, 2022, pp. 6749--6758.

\bibitem{pitteri2019object}
G.~Pitteri, M.~Ramamonjisoa, S.~Ilic, and V.~Lepetit, ``On object symmetries and 6d pose estimation from images,'' in \emph{2019 International conference on 3D vision (3DV)}.\hskip 1em plus 0.5em minus 0.4em\relax IEEE, 2019, pp. 614--622.

\bibitem{labbe2020cosypose}
Y.~Labb{\'e}, J.~Carpentier, M.~Aubry, and J.~Sivic, ``Cosypose: Consistent multi-view multi-object 6d pose estimation,'' in \emph{Computer Vision--ECCV 2020: 16th European Conference, Glasgow, UK, August 23--28, 2020, Proceedings, Part XVII 16}.\hskip 1em plus 0.5em minus 0.4em\relax Springer, 2020, pp. 574--591.

\bibitem{mo2022es6d}
N.~Mo, W.~Gan, N.~Yokoya, and S.~Chen, ``Es6d: A computation efficient and symmetry-aware 6d pose regression framework,'' in \emph{Proceedings of the IEEE/CVF Conference on Computer Vision and Pattern Recognition}, 2022, pp. 6718--6727.

\bibitem{richter2021handling}
J.~Richter-Klug and U.~Frese, ``Handling object symmetries in cnn-based pose estimation,'' in \emph{2021 IEEE International Conference on Robotics and Automation (ICRA)}.\hskip 1em plus 0.5em minus 0.4em\relax IEEE, 2021, pp. 13\,850--13\,856.

\bibitem{hodan2020epos}
T.~Hodan, D.~Barath, and J.~Matas, ``Epos: Estimating 6d pose of objects with symmetries,'' in \emph{Proceedings of the IEEE/CVF conference on computer vision and pattern recognition}, 2020, pp. 11\,703--11\,712.

\bibitem{hodan2024bop}
T.~Hodan, M.~Sundermeyer, Y.~Labbe, V.~N. Nguyen, G.~Wang, E.~Brachmann, B.~Drost, V.~Lepetit, C.~Rother, and J.~Matas, ``Bop challenge 2023 on detection, segmentation and pose estimation of seen and unseen rigid objects,'' \emph{arXiv preprint arXiv:2403.09799}, 2024.

\bibitem{zhou2019continuity}
Y.~Zhou, C.~Barnes, J.~Lu, J.~Yang, and H.~Li, ``On the continuity of rotation representations in neural networks,'' in \emph{Proceedings of the IEEE/CVF Conference on Computer Vision and Pattern Recognition}, 2019, pp. 5745--5753.

\bibitem{li2019cdpn}
Z.~Li, G.~Wang, and X.~Ji, ``Cdpn: Coordinates-based disentangled pose network for real-time rgb-based 6-dof object pose estimation,'' in \emph{Proceedings of the IEEE/CVF International Conference on Computer Vision}, 2019, pp. 7678--7687.

\bibitem{xiang2018posecnn}
Y.~Xiang, T.~Schmidt, V.~Narayanan, and D.~Fox, ``Posecnn: A convolutional neural network for 6d object pose estimation in cluttered scenes,'' 2018.

\bibitem{2017tless}
T.~Hodan, P.~Haluza, S.~Obdrzalek, J.~Matas, M.~Lourakis, and X.~Zabulis, ``T-less: An rgb-d dataset for 6d pose estimation of texture-less objects,'' \emph{IEEE}, 2017.

\bibitem{icbin}
A.~Doumanoglou, R.~Kouskouridas, S.~Malassiotis, and T.~Kim, ``Recovering 6d object pose and predicting next-best-view in the crowd,'' in \emph{2016 IEEE Conference on Computer Vision and Pattern Recognition (CVPR)}.\hskip 1em plus 0.5em minus 0.4em\relax Los Alamitos, CA, USA: IEEE Computer Society, jun 2016, pp. 3583--3592.

\bibitem{sundermeyer2018implicit}
M.~Sundermeyer, Z.-C. Marton, M.~Durner, M.~Brucker, and R.~Triebel, ``Implicit 3d orientation learning for 6d object detection from rgb images,'' in \emph{Proceedings of the european conference on computer vision (ECCV)}, 2018, pp. 699--715.

\bibitem{park2019pix2pose}
K.~Park, T.~Patten, and M.~Vincze, ``Pix2pose: Pixel-wise coordinate regression of objects for 6d pose estimation,'' in \emph{Proceedings of the IEEE/CVF International Conference on Computer Vision}, 2019, pp. 7668--7677.

\bibitem{wen2020edge}
Y.~Wen, H.~Pan, L.~Yang, and W.~Wang, ``Edge enhanced implicit orientation learning with geometric prior for 6d pose estimation,'' \emph{IEEE Robotics and Automation Letters}, vol.~5, no.~3, pp. 4931--4938, 2020.

\bibitem{fcos}
Z.~Tian, C.~Shen, H.~Chen, and T.~He, ``Fcos: Fully convolutional one-stage object detection,'' in \emph{Proceedings of the IEEE/CVF International Conference on Computer Vision (ICCV)}, October 2019.

\bibitem{shugurov2021dpodv2}
I.~Shugurov, S.~Zakharov, and S.~Ilic, ``Dpodv2: Dense correspondence-based 6 dof pose estimation,'' \emph{IEEE transactions on pattern analysis and machine intelligence}, vol.~44, no.~11, pp. 7417--7435, 2021.

\bibitem{castro2023crt}
P.~Castro and T.-K. Kim, ``Crt-6d: Fast 6d object pose estimation with cascaded refinement transformers,'' in \emph{Proceedings of the IEEE/CVF Winter Conference on Applications of Computer Vision}, 2023, pp. 5746--5755.

\end{thebibliography}

\end{document}